\documentclass{article}
\usepackage[final]{nips_2016}
\usepackage[utf8]{inputenc} 
\usepackage[T1]{fontenc}    
\usepackage{hyperref}       
\usepackage{url}            
\usepackage{booktabs}       
\usepackage{amsfonts}       
\usepackage{nicefrac}       
\usepackage{microtype}      
\usepackage{graphicx}
\title{Towards Wide Learning: Experiments in Healthcare}
\author{
  Snehasis Banerjee, Tanushyam Chattopadhyay, Swagata Biswas,\\
  \textbf{Rohan Banerjee, Anirban Dutta Choudhury and Arpan Pal}\\
  TCS Research and Innovation\\
  Tata Consultancy Services\\
  Kolkata, India 700160 \\
  \texttt{\{firstname.lastname\}@tcs.com} \\  
  \And
  Utpal Garain \\
  Indian Statistical Institute \\
  203, B.T. Road, Kolkata, India 700108\\
  \texttt{utpal@isical.ac.in} \\
}
\begin{document}
\maketitle
\begin{abstract}
In this paper, a Wide Learning architecture is proposed that attempts to automate the feature engineering portion of the machine learning (ML) pipeline. Feature engineering is widely considered as the most time consuming and expert knowledge demanding portion of any ML task. The proposed feature recommendation approach is tested on 3 healthcare datasets: a) PhysioNet Challenge 2016 dataset of phonocardiogram (PCG) signals, b) MIMIC II blood pressure classification dataset of photoplethysmogram (PPG) signals and c) an emotion classification dataset of PPG signals. While the proposed method beats the state of the art techniques for 2nd and 3rd dataset, it reaches 94.38\% of the accuracy level of the winner of PhysioNet Challenge 2016. In all cases, the effort to reach a satisfactory performance was drastically less (a few days) than manual feature engineering. 
 
\end{abstract}
\section{Introduction}
With the rapid growth in the availability and size of digital health data and wearable sensors, along with the rise of newer machine learning methods, health care analytics has become a hot area of research today.
The main bottlenecks for solving a healthcare data analytics problem are:
a) Effort required to build good models in terms of time, money and expertise
b) Interpreting model features so that a healthcare expert can do a causality analysis and take preventable measures or derive meaningful insights backed by domain knowledge.
A typical analytics solution requires a) Pre-processing b) Feature Extraction c) Feature Selection d) Modeling such as Classification or Regression. Among these steps, Feature Extraction and Feature Selection together form Feature Engineering (FE) and is the most time consuming and human expertise demanding among the rest.

Feature engineering can be broadly carried out in four ways: (a) manually selecting features guided by domain knowledge (b) recommending features by automated analysis - proposed method (c) feature transforms like Principal Component Analysis (PCA) (d) representation learning using deep architectures such as deep Multi-Layered Perceptron (MLP) and Convolutional Neural Network (CNN). Through experiments on 3 different types of healthcare datasets including a recent challenge dataset and comparison of the approaches, the utility of our proposed method (b) has been shown. Interpretation of features is not supported by deep learning and feature transform methods. But, manual feature engineering and our proposed method yield interpretable features which is very helpful in prognostic domains like healthcare.

\section{Solution Approach}
\begin{figure}[h]
\centering
\includegraphics[width=5.0in]{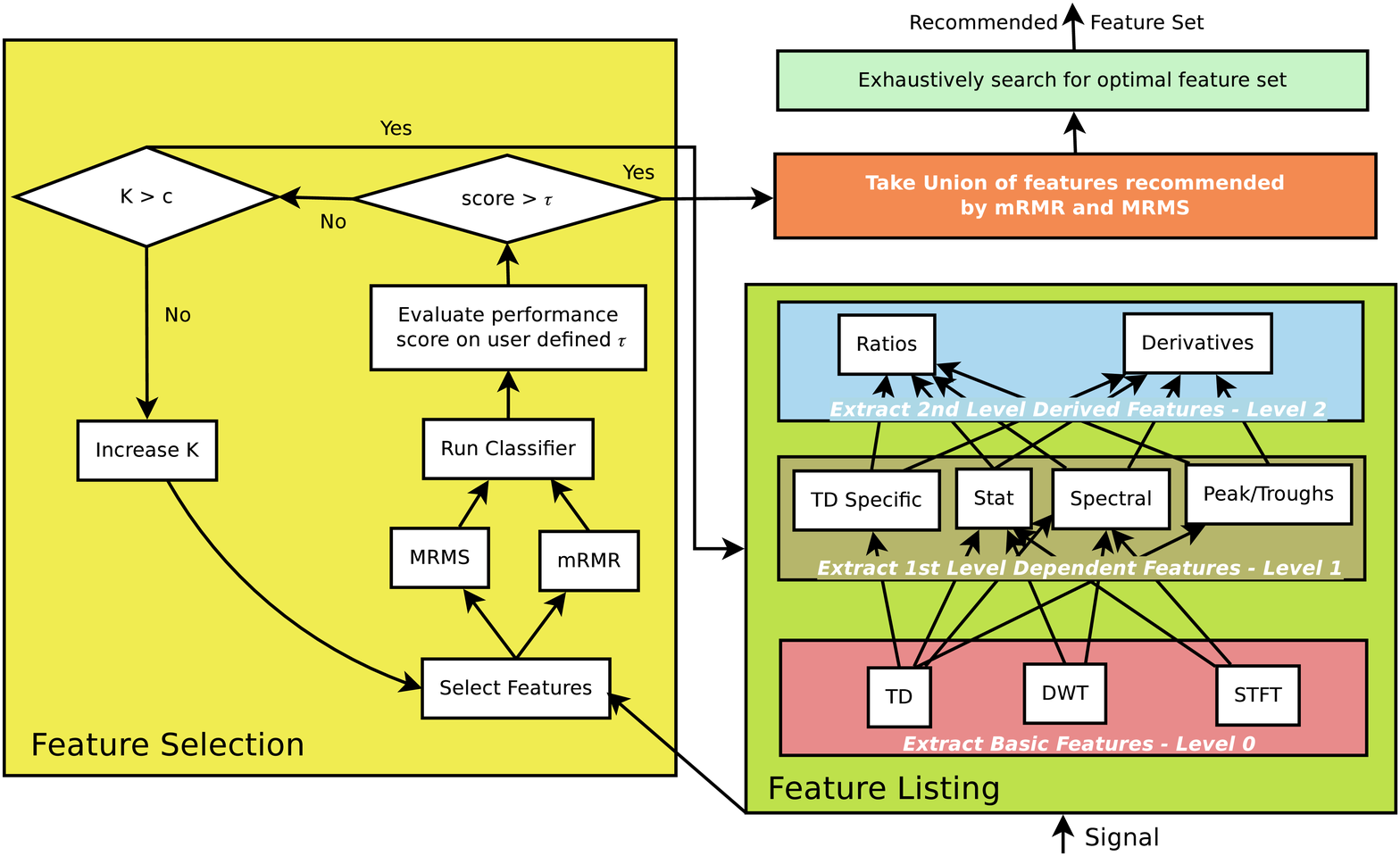}
\caption{Proposed Method of Feature Recommendation}
\label{fig:FeatureEngineering}
\end{figure}
While in Deep Architectures, the different activation functions can be hierarchically stacked to form new structures, in our approach, this does not hold true. For example, Wavelet transforms applied on Fourier transforms does not make sense. Hence, here the emphasis is on creating a Wide Architecture with meaningful hierarchies so that lowest layer contains basic feature extraction techniques, and as we move up we keep adding more meaningful layers on top of what was extracted. This helps in deriving physical interpretation of features (from bottom to top).
The dataset is partioned into p-folds of training, evaluation and testing sets (range of p is 5 to 10). The performance is reported on the hidden testing set. The proposed method consists of 3 steps: 

{\it 1. Feature Listing}: 
We have organized commonly reported features (in the literature of sensor data analytics) in a hierarchical manner as shown in Figure 1. The  basic features (level 0) can be mainly categorized as: (i) time domain features (TD) (ii) fourier transformation based features (FD) like short-time fourier transform (STFT) (iii) discrete wavelet transformation based features (DWT). One major challenge of using DWT features is the selection of suitable mother wavelet, as more than 100 different types of mother wavelets were reported in different papers. The automated mother wavelet selection is done by measuring energy to entropy ratio [1]. In level 1, spectral, statistical and peak-trough features are extracted. Level 2 includes different ratios and derivatives of the level 1 features. The system has capability of easy plugging of new feature extraction algorithms that will lead to a collaborative ecosystem. Hence, it is possible to get huge number (say, $N$) of features (including the transform domain coefficients) from the sensor signals. This results in $2^N-1$ possible combinations of features, whose exploration is practically infeasible, thereby demanding usage of feature selection.

{\it 2. Feature Selection}:
In our method, we followed an iterative feature selection where $k$-features are selected (k$\ll$N) at each iteration and system performance (e.g. classification accuracy) is checked for this feature set. If the selected feature set results in {\it expected} performance, we return the feature set as the recommended one. Otherwise, another $k$-features are chosen in the next iteration and the same steps are repeated. For checking the classification accuracy, we choose SVM (support vector machine) based classification with different kernels. SVM was selected as a classifier as it generalizes well and converges fast. Several values of $k$ are tried to choose an optimal value. For a given value of $k$, features are selected using two techniques namely, mRMR [2] and MRMS [3], described below:
{\it Minimum Redundancy and Maximum Relevance}(mRMR): In order to select effective features, mRMR optimizes an objective function, either Mutual Information Difference (MID) or Mutual Information Quotient (MIQ), by minimizing the redundancy and maximizing the relevance of the features. MID (additive) and MIQ (multiplicative) are defined as follows.\\
\centerline{$MID = \texttt{max} (V - W) \quad \quad, \quad \quad  MIQ = \texttt{max} (V/W)$} \\
where $V$ minimizes redundancy by computing F-statistics and $W$ maximizes relevance by computing correlation between a pair of features.\\
{\it Maximal Relevance Maximum Significance}$ $(MRMS): This technique uses fuzzy-rough set selection criteria to select relevant and non-redundant (significant) features. The objective function is:\\
\centerline{$J = J_{rel} + \beta J_{sig}$}\\
where $J_{rel}$ computes relevance of a recommended feature with respect to a class label and $J_{sig}$ computes the significance of a pair of recommended features by computing their correlation, and $\beta$ is the weight parameter.
Let, $x$ and $y$ be the sets of features recommended by mRMR and MRMS, respectively. Then the recommended set of features R is $z$, where $z = x \cup y$, where $|z|= k$. Note that mRMR and MRMS cover different aspects of feature selection. For instance, mRMR is classifier independent where as MRMS is effective to reduce real valued noisy features which are likely to occur in large feature sets.

{\it 3. Feature Recommendation}:
The system finds 2 feature sets for a particular performance metric (such as accuracy, sensitivity, specificity, precision, f-score): a) Fe1 - that produces the highest metric in any fold of cross-validation b) Fe2 - that is most consistent and performs well across all folds. The above step of feature selection is done hierarchically - if layer 0 does not produce expected results set by pre-set threshold $\tau$ or maximum possible value of a selected metric, then layer 1 is invoked. Similarly if layer 1 does not yield expected results, layer 2 is invoked. This follows the principle that if simple features can do the task, there is no need for complex features. `c' is a regularizer for `k' and is dependent on the hardware capabilities of the system. The intuition is that on a high-end machine (having higher valued `c'), feature combinations ($2^c-1$) can be carried in acceptable time. Using the recommended feature sets, any classifier like SVM or Random Forest can be trained to see the results obtained. Also by looking up the recommend features from the Feature Listing database, interpretation of the features can be easily obtained by a domain expert.
\section{Experiments and Results}

Experiments were carried on 3 datasets: D1, D2, D3 in order to provide a comparison among the feature engineering ways (proposed method, manual, dimension reduction and deep learning).

{\it D1:} The Physionet 2016 Challenge dataset [4] consists of 3153 heart sounds, including 2488 normal and 665 abnormal recordings. The ground truth label (normal or abnormal heart sound) of each record is manually annotated by expert doctors. Raw PCG (phonocardiogram) is further down sampled to 1 KHz from 2 KHz, in order to segregate four cardiac states (S1, systole, S2 and diastole) using the logistic regression based HSMM approach [5]. The winner [6] of the challenge used 124 features and used deep learning for classification. The challenge used their own modified metric for ranking participants, however for consistency of results across datasets, we have used accuracy score as the performance metric. We participated in the challenge using manual features and got only 1\% increase in performance compared to the proposed automated method. 

{\it D2:} The second dataset is derived from MIMIC-II patients dataset [7]. A subset of the dataset containing PPG (photoplethysmogram) data was created after noise cleaning and the ground truth blood pressure (BP) was obtained from the simultaneously recorded arterial BP waveform, resulting in equally balanced 36 high (>140 mmHg reading) and 36 low BP patient waveform data instances.

{\it D3:} The third dataset (used to classify the emotion into happy and sad) records the fingertip pulse oximeter PPG data of 33 healthy subjects (Female: 13 and Male: 20) with average age 27 years. We used standard video stimuli as ground-truth and time synchronization errors were minimized.

Table 1 lists the obtained result for a dataset along with the corresponding configuration and effort for each of the feature engineering approaches. 
Experiments has been carried out using Theano \footnote{Theano version 0.8.2 used from http://deeplearning.net/software/theano/}  based Multi-Layer Perceptron with \textit{Dropout} and varying number of layers to see if features can be automatically learned on the datasets under experimentation. Different epochs (5 to 15) has been tried to see how the learning rate affects performance. Different activation functions like rectified linear unit (relu), tanh, softmax, sigmoid, etc. has been tried out at different layer level to get an ideal architecture for classification task for the given problems. Table 1 shows that MLP based techniques fail when compared to the state of the art and the proposed method. The problem with MLP and newer deep learning techniques like CNN is that they need a lot of data to train and there is no way to interpret the features.
Principal component analysis (PCA) is a statistical procedure that uses an orthogonal transformation to derive principal components representative of the features under consideration. Experiments have been carried out with aforementioned datasets and Gaussian kernel is used for SVM based classification. The different dimension reduction techniques used are Singular Value Decomposition (svd), Eigen Value Decomposition (eig) and Alternating Least Squares (als). A varying number of principal components (like 5, 10, 15) are also tried out. Table 1 shows that PCA based methods are outperformed by our proposed method. Another drawback of PCA and similar feature reduction techniques is that the derived features are not interpretable. It is seen that for 2nd and 3rd dataset, the proposed approach outperforms the state of the art (SoA) methods, and for the 1st dataset, 94.38\% of the accuracy level of the winner was reached by this method. In terms of effort taken to build the solution, the proposed method clearly beats others.
\section{Conclusion and Future Work}
\begin{table*}[t]
\caption{Comparative study of the 4 techniques (MLP, PCA, manual SoA, proposed WIDE method)}
\label{Table:comparison}
\centering
\begin{tabular}{l|c|c|c|c|c|}
\hline
Datasets (D) & MLP Acc. & PCA+SVM Acc. & SoA Acc. & WIDE Acc.\\
\hline
D1. PhysioNet Challenge & 0.79 (relu)  & 0.83 (svd, 5 comp.) & 0.89 $^W$ , 0.85 $^U$ & 0.84\\
\hline
D2. MIMIC II BP & 0.5 (relu) & 0.625 (eig, 5 comp.) & 0.795 & 0.878 \\
\hline
D3. Emotion (in house)& 0.5 (tanh) & 0.5 (svd, 10 comp.) & 0.823 & 0.909 \\
\hline
Effort in person-days & D1. 10 & D1. 5 & D1. 120 & D1. 3 \\
unit (from raw dataset & D2. 5 & D2. 5 & D2. 90 & D2. 2 \\
to results and analysis) & D3. 7  & D3. 3  & D3. 130  & D3. 2 \\
\hline
Interpretable features & No & No & Yes & Yes \\
\hline
\end{tabular}
\begin{tabular}{l}
\small W = score of winner; U = score of our team that participated using manual features
\end{tabular}
\end{table*}
Interpretable Feature Engineering has been found to be the most demanding task among all the subtasks of health data analytics. Hence, a system was built to automate this part of the process. The system has been tested on three healthcare datasets and was found to give good results when compared to state of the art. Apart from manual feature engineering, comparison has been made with MLP and PCA which are feature engineering approaches of different directions. Interpretation of features is one of the strong points of the proposed method. Another strong point of the proposed method is huge reduction in effort to develop a typical analytics solution. Integration of knowledge bases for ease of interpreting features and automated causality analysis is also planned. The work will be exteneded to other domains such as machine prognostics.

\section*{References}
\small

[1] Ngui, W. K. et al. (2013) Wavelet Analysis: Mother Wavelet Selection Methods, {\it Applied Mechanics and Materials}; Vol. 393, pp. 953-958.

[2] Peng, H et al. (2005) Feature selection based on mutual information criteria of max-dependency, max-relevance, and min-redundancy. {\it IEEE Transactions on Pattern Analysis and Machine Intelligence}, 27.8: 1226-1238.

[3] Maji, P et al (2012) Fuzzy-Rough MRMS Method for Relevant and Significant Attribute Selection, {\it Advances on Computational Intelligence: 14th International Conference on Information Processing and Management of Uncertainty in Knowledge-Based Systems.}

[4] Liu, C et al. (2016) An open access database for the evaluation of heart sound algorithms. {\it Physiological Measurement}; 37(9).

[5] Springer et al. (2016) Logistic regression-hsmm-based heart sound segmentation. {\it IEEE Transactions on Biomedical Engineering};63(4):pages 822–832.

[6] Potes, C et al. (2016), {\it Ensemble of Feature-based and Deep learning-based Classifiers for
Detection of Abnormal Heart Sounds}, CiNC 2016.

[7] Goldberger, AL et al. (2000) PhysioBank, PhysioToolkit, and PhysioNet: Components of a New Research Resource for Complex Physiologic Signals. {\it Circulation} 101(23):e215-e220.

\end{document}